\documentclass[lettersize,journal]{IEEEtran}
\usepackage{amsmath,amsfonts}
\usepackage{algorithmic}
\usepackage{algorithm}
\usepackage{array}
\usepackage[caption=false,font=normalsize,labelfont=sf,textfont=sf]{subfig}
\usepackage{textcomp}
\usepackage{stfloats}
\usepackage{url}
\usepackage{verbatim}
\usepackage{graphicx}
\usepackage{multirow}
\usepackage[style=ieee, sorting=none]{biblatex}
\begin{document}

\title{High-Fidelity Digital Twins for Bridging the Sim2Real Gap in LiDAR-Based ITS Perception}

\author{Muhammad Shahbaz, Shaurya Agarwal (\IEEEmembership{Member,IEEE})

\thanks{This paper was produced by the IEEE Publication Technology Group. They are in Piscataway, NJ.}
\thanks{Manuscript received April 19, 2021; revised August 16, 2021.}
\thanks{Muhammad Shahbaz and Shauray Agarwal are with the Department of Civil, Environmental and Construction Engineering, University of Central Florida, Orlando, FL 32816, USA.}}

\markboth{Journal of \LaTeX\ Class Files,~Vol.~14, No.~8, August~2021}%
{Shell \MakeLowercase{\textit{et al.}}: A Sample Article Using IEEEtran.cls for IEEE Journals}

\IEEEpubid{0000--0000/00\$00.00~\copyright~2021 IEEE}

\maketitle

\begin{abstract}
Sim2Real domain transfer offers a cost-effective and scalable approach for developing LiDAR-based perception (e.g., object detection, tracking, segmentation) in Intelligent Transportation Systems (ITS). However, perception models trained in simulation often under perform on real-world data due to distributional shifts. To address this Sim2Real gap, this paper proposes a high-fidelity digital twin (HiFi DT) framework that incorporates real-world background geometry, lane-level road topology, and sensor-specific specifications and placement.
We formalize the domain adaptation challenge underlying Sim2Real learning and present a systematic method for constructing simulation environments that yield in-domain synthetic data. An off-the-shelf 3D object detector is trained on HiFi DT–generated synthetic data and evaluated on real data. Our experiments show that the DT-trained model outperforms the equivalent model trained on real data by 4.8\%. To understand this gain, we quantify distributional alignment between synthetic and real data using multiple metrics, inclding Chamfer Distance (CD), Maximum Mean Discrepancy (MMD), Earth Mover’s Distance (EMD), and Fréchet Distance (FD), at both raw-input and latent-feature levels.
Results demonstrate that HiFi DTs substantially reduce domain shift and improve generalization across diverse evaluation scenarios. These findings underscore the significant role of digital twins in enabling reliable, simulation-based LiDAR perception for real-world ITS applications.
\end{abstract}

\begin{IEEEkeywords}
Digital Twin, CARLA Simulation, Simulation Fidelity,
Sim2Real, Domain Transfer, Lidar Perception, 3D Object Detection, Feature Analysis
\end{IEEEkeywords}

\section{INTRODUCTION}

\noindent \textbf{Context and Motivation:} The light detection and ranging (LIDAR) technology offers robust performance in terms of accuracy and light agnostic operation. It is becoming foundational in advancing perception algorithms for intelligent transportation systems (ITS). These perception algorithms perform tasks such as 3D object detection, tracking, and semantic and instance segmentation. To develop lidar-based perception algorithms, deep neural networks remain at the forefront. However, most of the methods to train those networks are based on supervised learning, that is, they need labeled data to be able to learn features from raw point-cloud data and generate tasks-specific predictions. For example, labels containing bounding box information (location, size, and orientation) is required, for each object of interest (bike, car, truck, etc.) in each point cloud frame, for training deep 3D object detectors. To achieve higher accuracy and generalization, the deep neural network models require large amounts of labeled data for training and testing of deep perception models \cite{zhai2024phase}. Towards that need, numerous high-quality lidar datasets has been curated such as KITTI \cite{geiger2013vision}, Waymo Open \cite{sun2020scalability}, NuScenes \cite{caesar2020nuscenes}, and others \cite{chang2019argoverse, geyer2020a2d2, patil2019h3d, mao2021one, xiao2021pandaset} for autonomous driving, IPS300+ \cite{wang2022ips300+}, LUMPI \cite{busch2022lumpi}, TUMTraf-I \cite{zimmer2023tumtraf}, and others \cite{mirlach2025r, cress2022a9, ye2022rope3d} for roadside lidar-based perception, and DAIR-V2X \cite{yu2022dair}, V2X-Real \cite{xiang2024v2x}, BAAI-VANJEE \cite{yongqiang2021baai}, and others \cite{xu2023v2v4real, krajewski2018highd} for collaborative perception tasks. However, creation and expansion of such datasets remains a challenge as: (1) recording and labeling a large number of point-cloud frames can incur high cost in terms of time, effort, and money; (2) human annotations can be biased and/or contain errors degrading label quality; (3) local factors such as road-user types (bikes, cars, trucks, etc.), environmental conditions (weather and lighting), etc. differ a lot among different geographical regions making data from one location less applicable to others; (4) extreme case scenarios such as traffic accidents are hard to collect; and (5) any changes to sensor characteristics (specifications and/or installation settings) shifts the data distribution, due to the factors such as increased resolution, point rate, point-of-view etc. Therefore, there is a significant need to develop cost-effective methods that can rapidly create high-quality lidar datasets containing close-to-real data distributions, and are highly adaptable in terms of specific scenario generation and changes in sensor characteristics.

\vspace{1mm}
\noindent \textbf{Sim2Real for ITS:} The idea of simulation-to-real training is a promising path towards economical and scalable perception systems. For lidar-based ITS applications, generating labeled data using a simulation environment typically involves a physics simulation of actors in a 3D environment (e.g., traffic at an intersection). The actor types can be primarily divided into two categories, static actors (surrounding buildings, trees, roads, etc.) and dynamic actors (pedestrians, bikes, cars, trucks, etc.). The synthetic lidar and associated label data is generated by emulating sensor(s) in the simulation of these 3D actors. This data should ideally resemble the target domain data, captured using real sensor(s) from real-world traffic scene(s), for effective Sim2Real learning.

\noindent \textbf{Limitations of Simulators:} To create realistic simulation advanced simulators are required. In \cite{li2024choose}, the authors comprehensively discuss numerous simulation tools for ITS, including traffic and sensory simulator. Among them, the comprehensive open-source simulators such as CARLA \cite{dosovitskiy2017carla}, DeepDrive \cite{team2020deepdrive}, and Vista \cite{amini2022vista} support a range of perception, planning, and control tasks for autonomous driving systems (ADS). Though their goal is to simulate autonomous driving scenarios, the 

\newpage \noindent functionality can often also be extended to support tasks for roadside and collaborative lidar-based perception applications.

The key benefit of using simulators is unlimited data generation, allowing users to create custom scenes, scenarios, traffic, and sensor configurations at a negligible cost. However, the usefulness of a dataset is not defined by the quantity alone, but also by the quality in terms of its faithfulness to target tasks. The faithfulness is usually measured by realism, distribution, and diversity of a dataset \cite{nowruzi2019much}. The built-in assets in the simulators are usually hand-crafted and are not realistic enough, resulting in data that does not conform to real point-cloud distributions. This introduces a significant sim-to-real gap \cite{manivasagam2020lidarsim}. As result, models trained on such synthetic data often exhibit poor performance in real-world deployments \cite{8864642}.

\noindent \textbf{Bridging Sim2Real Gap using HiFi DTs:} For effective Sim2Real learning, it is important that the synthetic and real data share a common feature space and distribution \cite{triess2021survey}. However, creating sensor simulations that align to real-world data is a highly challenging task, requiring a faithful representation of both static and dynamic aspects of the target reality in simulation. Towards bridging this gap, this paper explores the applicability of high-fidelity digital twins (HiFi DTs) in training lidar-based perception models. Particularly, a 3D object detection tasks is studied, and the HiFi DT is employed to create the training set for that purpose. By incorporating both the static geometric layout (surrounding buildings, trees, roads, etc.) and the dynamic context (traffic types and flow, and sensor specifications and locations) in HiFi DT simulation, the data is aligned to the target reality. To formalize the core challenge addressed in this work, we consider the distributional divergence between synthetic and real-world data in the context of supervised learning. By minimizing this domain gap using HiFi DTs, our goal is to ensure that models trained in simulation generalize well to real-world settings. A detailed mathematical formulation of this objective is presented in Section~\ref{subsec:formulation}.

\begin{figure*}[ht]
  \centering
  \begin{minipage}[t]{0.98\textwidth}
    \includegraphics[width=\textwidth]{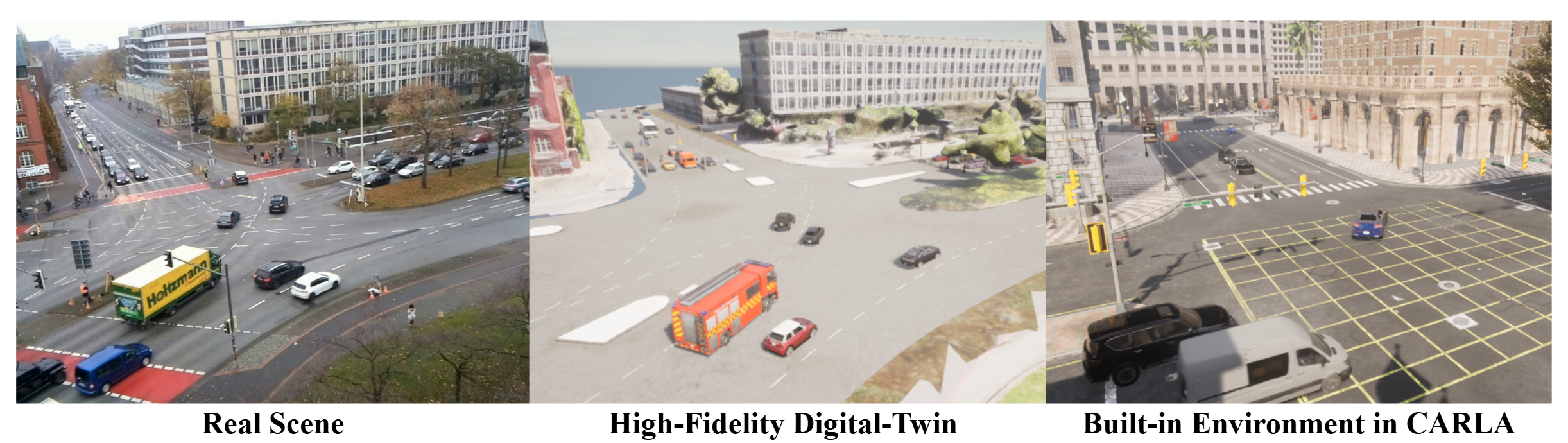}
    \caption{A qualitative comparison between real scene, our high-fidelity digital twin, and generic built-in CARLA environment. In our modeling, numerous factors such as surrounding buildings, in-scene vegetation, and lane-level road topology and alignment is carefully constructed to achieve a close-to-real simulation. Naive CARLA simulations lack such fidelity.}
    \label{fig:orig_vs_dt_synth}
  \end{minipage}
\end{figure*}

Our approach exploits publicly available map imagery (2D and 3D images), OpenStreetMap \cite{OpenStreetMap} information, and lane-level data (location, height, and superelevation) to construct the background geometry and road. Then, traffic is generated, however stochastically, to match statistical distribution of the object classes in the region of interest. And finally, synthetic point cloud dataset along 3D bounding-box labels is generated using sensor simulation adhering to real sensor(s) specifications (resolution, point rate, range, etc.) and placement details (location and orientation). Our method results in highly faithful simulation, see Fig.~\ref{fig:orig_vs_dt_synth}. The generated dataset is then used to train an off-the-shelf model (SEED \cite{liu2024seed}), and the model's accuracy is compared against an identical model, trained on real data. The HiFi-DT-trained model outperformed its real-data-trained opponent by 4.8\%, setting a new standard for effective Sim2Real learning.

To establish that the generated data is in-domain to real data, the paper further dives deep into input-data and latent-feature level analysis. First, statistical distribution analysis is conducted noting mean- point densities, class distributions, and object sizes and counts. Additionally, Chamfer Distance is calculated to quantify geometric similarity. Later, nonlinear dimensionality reduction techniques, t-SNE and UMAP \cite{mcinnes2018umap}, are employed to plot latent-feature similarity (from SEED's backbone) for synthetic and real data. Finally, two domain-discrepancy metrics, Maximum Mean Discrepancy (MMD) and Earth Mover's Distance (EMD) are computed to showcase alignment. The high similarity trends demonstrate reduced Sim2Real gap. Moreover, a high overlap in latent-features, generated by detector's backbone, demonstrate that synthetic and real data exhibit similar distributions in feature space, suggesting that the model perceive them as structurally and semantically alike. 

For evaluations, an established real-world benchmark, the Leibniz University Multi-perspective Intersection (LUMPI) \cite{busch2022lumpi} dataset is used throughout the course of this article. It is a large-scale roadside lidar dataset, captured using 5 lidar sensors, from Königsworther Platz intersection, located in Hannover, Germany. Since the original dataset contains 90K+ frames, a subset \textit{Measurement4} of the dataset is used to keep the experimentation feasible. The complete synthetic dataset used for training, \texttt{UT-LUMPI}\cite{ucf_ut_lumpi}, is available at our Harvard dataverse \url{https://dataverse.harvard.edu/dataverse/ucf-ut}. Researchers are encouraged to utilize our other HiFi DT-based datasets, including \texttt{UT-V2X-Real} \cite{ucf_ut_v2x_real_ic} and \texttt{UT-TUMTraf-I} \cite{ucf_ut_tumtraf_i}, modeled after original V2X-Real \cite{xiang2024v2x} and TUMTraf-I \cite{zimmer2023tumtraf} dataset.

\subsection{Contributions}

In summary, the key {contributions} of this paper are as follows:
\begin{itemize}
\item To the best of our knowledge, this is the first study that employs HiFi DT, that takes into account both static and dynamic aspects of real scene, to generate lidar data that is directly applicable for training lidar-based perception models for real world.
\item It develops a feasible and systematic method to create high-fidelity digital twins of real locations using publicly available information.
\item It presents a technique to accurately model lidar-sensor simulation on top of HiFi DTs.
\item It trains, for the first time, a deep 3D object detector on HiFi-DT-based (synthetic) dataset that outperforms identical model trained on real data.
\item An in-depth distribution analysis, showing reduced Sim2Real gap, is conducted for point cloud data at both raw-input and latent-feature levels of the deep detector model, utilizing multiple established metrics.
\end{itemize}

Before presenting our methodology, we first review related efforts in simulation, domain adaptation, and digital twin–based perception modeling.

\section{Related Work and Gaps}
It is established from years of research that unequal sources lead to performance discrepancies \cite{ben2010theory}. In the context of this paper, studies \cite{wang2020train, tsai2022see} tell that models trained on real data at one location underperform on data at an other location. Sim2Real learning offers solution, however naive implementation results in huge performance degrades. For example, Dworak et al. \cite{dworak2019performance} tested CARLA-simulation trained object detectors on KITTI \cite{geiger2013vision} dataset, and noted a 68\% performance decline in mAP scores. Such decline is also noted by Fang et al. \cite{fang2020augmented} and Manivasagam et al. \cite{manivasagam2020lidarsim} in their studies. It is important to note that object detection is just one type of perception tasks, studies such as by Yue et al. \cite{yue2018lidar} and Spiegel et al. \cite{spiegel2021using} focus on semantic segmentation instead, and present that this gap still emerges. To counter that, multiple research paths are studied including improved sensor modeling, domain adaptation, and digital-twin modeling.

\vspace{1mm}
\noindent \textbf{Sensor Simulation Studies:} Comprehensive simulators, such as CARLA \cite{dosovitskiy2017carla}, offer numerous features for autonomous driving applications including lidar sensors. However, their sensor models lack many real-world effects \cite{guillard2022learning}. Simulators such as LGSVL Simulator \cite{rong2020lgsvl} put more emphases on sensor fidelity. However, still effects such as meterial reflectance, ray dropout on glass, multiple returns, or motion distortions, etc. are not modeled. In \cite{zhou2023garchingsim} GarchingSim is presented, that generates photorealistic scenes for autonomous driving. Haider et al. \cite{Haider2022HighFidelity} studied accurate ray-tracing to accurately model lidar. They developed a complete signal-processing pipeline and tested it against real-world measurements. They showed that sensor imperfections, such as optical losses, electronic noise, and multi-echo behavior, must be accurately modeled to increase simulation fidelity. Recent research by Manivasagam et al. \cite{manivasagam2020lidarsim} blends accurate ray casting to mimic sensors noise producing realistic point clouds than pure CAD models \cite{manivasagam2020lidarsim}. These studies demonstrate an increased fidelity in sensor simulations. However, it is important to note that accurate sensor physics alone, in an otherwise made-up simulation scene, is not sufficient to account for geometrical shapes and dynamics seen in real data.

\vspace{1mm}
\noindent \textbf{Domain Adaptation Studies:} Generally, domain adaptation methods focus on transferring trained model from one domain to another. In ITS, an early work is SqueezeSeg \cite{wu2018squeezeseg}, that trained a lidar segmentation network on synthetic data and fine-tuned on real scans. Its improved version, SqueezeSegV2 \cite{wu2019squeezesegv2}, introduced an unsupervised domain adaptation pipeline that learns an intensity mapping and aligns feature distributions to mitigate the simulation-to-real gap. They reported a 2x improvement in segmentation accuracy. These studies sparked later research to bridge domain differences for lidar-based perception tasks. Zhao et al. \cite{zhao2021epointda} proposed an end-to-end framework, ePointDA, that injects dropout noise into synthetic point cloud data to mimic missing points. Additionally, it performs spatially adaptive feature alignment, and avoids prior knowledge of real data statistics. It attained, trained on GTA-V simulator, superior segmentation on real SemnaticKITTI \cite{behley2019semantickitti}. In \cite{Haghighi2024CoLiGen}, authors introduced CoLiGen, a framework that generatively converts lidar data to depth-reflectance images, using GANs to translate simulated scans into more realistic point clouds. Yang et al. developed ST3D \cite{yang2021st3d}, that iteratively refines a detector trained on a source domain using target-domain pseudo-labels and curriculum augmentation. They showed that carefully designed adaptation, such as scaling object sizes to reduce source bias, then filtering and retraining with target pseudo-labels, can yield robust cross-domain detectors.

Domain adaptation methods focus transferability of data post generation. The need for such corrective strategies highlights the underlying issue: the simulation domain itself is deficient compared to reality. By contrast, digital-twin modeling focus on minimizing that initial gap. If the simulator can produce lidar data that is almost indistinguishable from real sensor data, the burden on feature-level domain adaptation can be greatly reduced.

\vspace{1mm}
\noindent \textbf{Digital-Twin Studies:} The idea of digital-twin-assisted simulation modeling for  perception tasks is quite recent. Li et al. \cite{li2025digital} constructed a high-fidelity digital twin for the first time to generate a labeled synthetic image dataset. They further applied a graph-matching module to align feature distributions between the simulated and real domains. By combining this along a multi-task network (object detection and segmentation) and employing multi-level domain discriminators, transferable representations are learned. In \cite{strunz2024cross}, researchers presented a realistic synthetic lidar dataset particularly focused on replicating the geometric and distributional properties of the KITTI and nuScenes datasets. They found, through cross-dataset generalization experiments, that the  models trained on their LUCID dataset achieved the best cross-dataset generalization performance, a mean cross-dataset average precision (AP-CDP) delta of -0.18 and average recall (AR-CDP) delta of -0.16, outperforming other real and synthetic datasets.

This digital twin-assisted simulation approach demonstrated an improved performance compared to isolated improvements of sensor physics and to prior UDA methods, highlighting its significance. However, digital-twin modeling can be a sophisticated process, particularly, because of all the digital asset (such as buildings, roads, road users, etc.) required to construct the map.

\vspace{1mm}
\noindent \textbf{Research Gaps and Motivation}: Despite the above progress, several key gaps remain in Sim2Real lidar learning: (1) Though, simulators incorporate improved lidar-sensor physics and noise models, their built-in environments are completely hand-crafted and are in most cases too much simplified to be used for Sim2Real learning; (2) Domain adaptation methods are primarily post-hoc corrective solutions, requiring further processing and feature-level domain adaptions; and (3) although HiFi DT modeling has emerged for lidar-based simulations, however, the process to create HiFi DTs is not streamlined. Our paper, inspired by exciting research in the domain, introduces a novel much simpler systematic method to create HiFi DTs for arbitrarily any location, using only public information. These HiFi DTs become basis for CARLA-based simulation that takes into account accurate sensor(s) pose and specification to create almost indistinguishable point cloud data. By adding simple gaussian noise, the data can be made more generalized to be used for target real location. Since resulting point clouds are, from the base, aligned to real locations, it is not required to do post processing. This makes our method much more useful compared to prior methods for Sim2Real learning.

\section{Problem Formulation and Methodology} \label{subsec:formulation}
\subsection{Problem Formulation} 

Let $\mathcal{D}_{\text{real}} = \{(x_i^{\text{real}}, y_i^{\text{real}})\}_{i=1}^{N}$ denote a real-world dataset (e.g., LUMPI), where $x_i^{\text{real}}$ are input point clouds and $y_i^{\text{real}}$ are the corresponding 3D object detection labels. These samples are drawn from a real-world joint distribution $P_{\text{real}}(x, y)$. Likewise, let $\mathcal{D}_{\text{sim}} = \{(x_j^{\text{sim}}, y_j^{\text{sim}})\}_{j=1}^{M}$ represent the synthetic dataset generated via a high-fidelity digital twin (HiFi DT), drawn from a simulation-based distribution $P_{\text{sim}}(x, y)$.

Our objective is to train a deep 3D object detection model $f_\theta : \mathcal{X} \rightarrow \mathcal{Y}$, parameterized by $\theta$, using $\mathcal{D}_{\text{sim}}$, such that it generalizes well on unseen real-world data. The goal is to minimize the expected detection loss over the real-world distribution:

\begin{equation}
\mathcal{E}_{\text{real}}(\theta) = \mathbb{E}_{(x, y) \sim P_{\text{real}}} \left[ \mathcal{L}(f_\theta(x), y) \right],
\end{equation}

where $\mathcal{L}$ is the composite loss function comprising classification, bounding box regression, and IoU-based components.

However, since $f_\theta$ is optimized on $\mathcal{D}_{\text{sim}}$, the generalization error $\mathcal{E}_{\text{real}}(\theta)$ depends on the divergence between the simulated and real-world data distributions:

\begin{equation}
\Delta(P_{\text{sim}}, P_{\text{real}}) = \text{divergence}(P_{\text{sim}}, P_{\text{real}}).
\end{equation}

Our central hypothesis is that by leveraging high-fidelity digital twins that faithfully replicate real-world geometry, sensor placement, and traffic patterns, we can construct $P_{\text{sim}}$ such that:

\begin{equation}
\Delta(P_{\text{sim}}, P_{\text{real}}) \rightarrow 0,
\end{equation}

thereby ensuring that:

\begin{equation}
\mathcal{E}_{\text{real}}(\theta) \approx \mathbb{E}_{(x, y) \sim P_{\text{sim}}} \left[ \mathcal{L}(f_\theta(x), y) \right].
\end{equation}

In this work, we empirically estimate the distributional alignment using both geometric and latent-feature metrics. Specifically, we compute Chamfer Distance (CD) in the point cloud space, and Maximum Mean Discrepancy (MMD), Earth Mover’s Distance (EMD), and Fréchet Distance (FD) in the latent feature space of the detector backbone. For example, the MMD is computed as:

\begin{equation}
\Delta_{\text{MMD}} = \left\| \mathbb{E}_{x \sim P_{\text{sim}}}[\phi(x)] - \mathbb{E}_{x \sim P_{\text{real}}}[\phi(x)] \right\|^2,
\end{equation}

where $\phi(x)$ denotes the embedding of input $x$ in the detector's latent feature space. A low value of $\Delta_{\text{MMD}}$ indicates strong feature-level alignment, suggesting that the detector perceives both domains as structurally and semantically similar.

This formulation serves as the core premise of our work: that high-quality synthetic data, when generated via scene- and sensor-faithful digital twins, can effectively minimize the Sim2Real gap and enable strong generalization of deep perception models in real-world ITS deployments.


\begin{figure*}[ht]
\centerline{\includegraphics[width=0.98\textwidth]{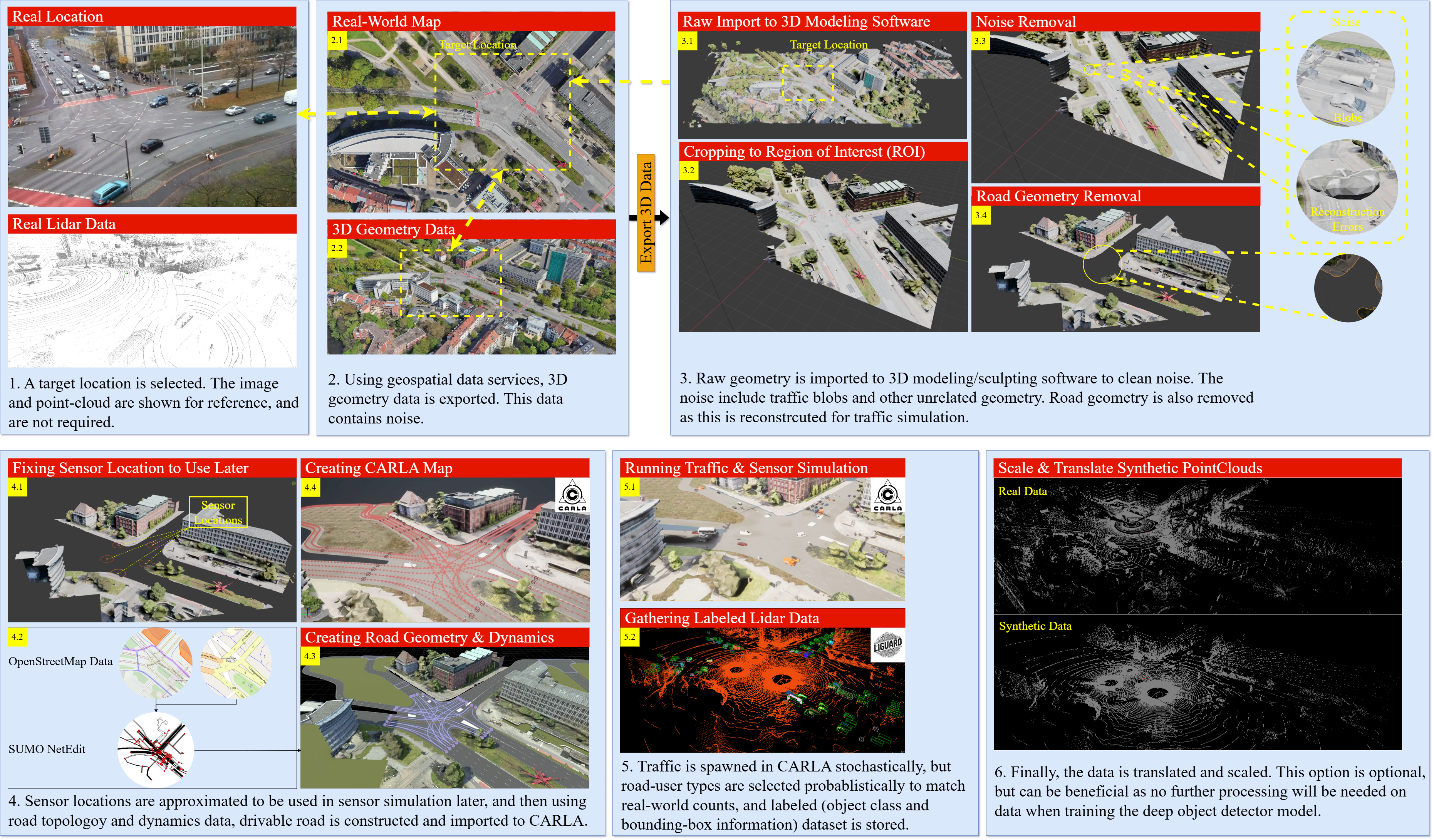}}
\caption{Systematic Method to Model High-Fidelity Digital Twin for Real-World Locations.
\label{fig:dt_modeling}}
\end{figure*}

\begin{figure*}[hb]
\centerline{\includegraphics[width=0.98\textwidth]{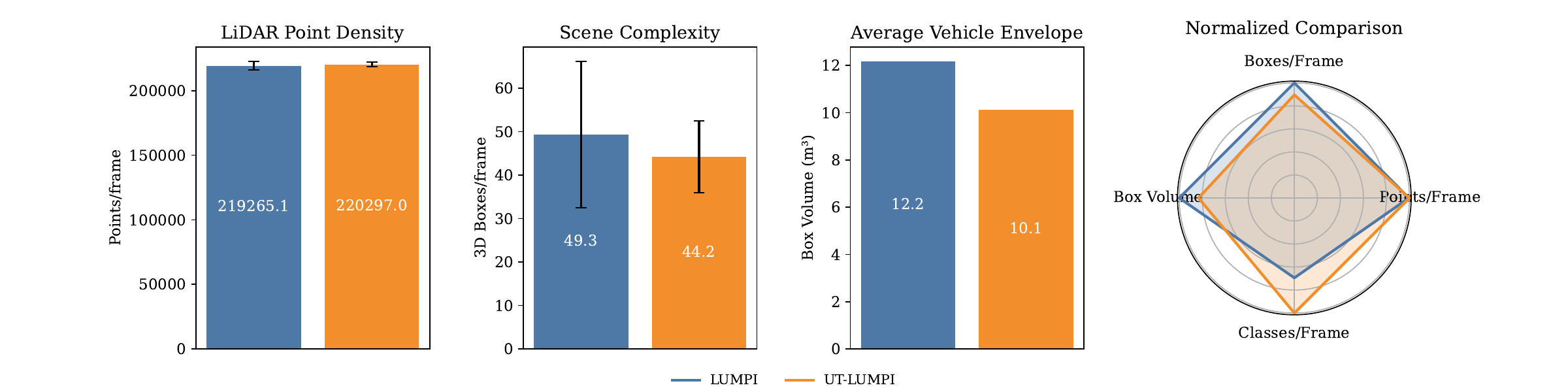}}
\caption{Frame-Level Statistical Comparison of Synthetic and Real Lidar Datasets: The alignment in point density emerges due to accurate sensor modeling. The scene complexity and bounding-box volumes are a function of stochastic road-user spawning. A high overlap in normalized comparison demonstrate a high similarity between synthetic and real dataset.
\label{fig:ut_lumpi_v_lumpi}}
\end{figure*}

\subsection{Synthesizing Close-to-Real Data Using HiFi DTs}
To create a realistic labeled lidar dataset, a step-by-step method methodology is as follows: (1) a target location is identified, and (2) 3D image of the target location is gathered using satellite map services such as Cesium 3D Tiles. The satellite 3D image is often a 3D mesh reconstruction, containing an estimation of the structures in the target reality. It contains low-poly geometries, meaning that the surfaces in the 3D mesh are not detailed. However, since real lidar data is also inherently sparse, therefore low-level surface details are not important in this use case. Another limitation of a satellite-based 3D image is that, often times it can contain traffic blobs because the source 2D images, that were used to reconstruct mesh, had such noise present. Another source of noise can be the reconstruction algorithm itself as it not perfect and can introduce noise like adding extra geometry that is not present in the real scene. Moreover, some geometry data outside the region of interest can also be still present in the captured 3D image and needs removal for efficiency reasons. Fortunately, removing such noise is rather easy using any 3D modeling tool such as Blender. So, (3) by loading the entire 3D mesh in 3D modeling tool, the mesh can is cut to region of interest, followed by removing extra geometry (traffic blobs, 3D shards in mid air, etc.) based on qualitative inspection. Additionally, the model is rescaled to match real-world units. Optionally, for convenience later, sensor positions are also noted at this stage. Once the 3D model is ready, the next step (4) is to create accurate roads that can be used by simulation tools to spawn traffic. In order to create accurate road topology inside region of interest, the road topology can be imported from either map tools such as OpenStreetMaps \cite{OpenStreetMap}, etc. CARLA supports multiple methods to create roads such as SUMO NetEdit \cite{SUMO2018} (open source), RoadRunner \cite{RoadRunner} (premium), or using CARLA's source project to create roads. Based on the preference any supported toolset can be utilized. Once the road topology is created and the resulting mesh is built, both the satellite-based mesh and road are imported into CARLA and are translated and/or rotated to fit each other. This completes the static part of digital twin. Next, (5) utilizing CARLA's original road-user catalog, traffic is spawn. The road-user types and distributions can be programmed into the traffic spawn script to match, statistically, the distribution of target real location. Finally, the data acquisition script, incorporating real sensor specifications and pose is executed to gather point-cloud and bounding-box (label) data. In our HiFi DT, the sensor specification include number of channels, horizontal resolution, horizontal and vertical fields of view, measurement range, and point rate. For sensor pose, center and tilt are considered. The synthetic dataset is stored in OpenPCDet \cite{openpcdet2020} format. An overview of the entire process is summarized in Fig.~\ref{fig:dt_modeling}.

Intuitively, due to digital-twin modeling, the resulting simulation closely approximates the reality, as the 3D spatial representation and traffic dynamics exhibit a high degree of correspondence to those observed in the real world, see Fig.~\ref{fig:ut_lumpi_v_lumpi}. This approach offers a cost-effective and computationally efficient approximation, providing sufficient fidelity for the intended applications. Please note that HiFi DT's specifications can either be aligned to existing data for augmentation purposes or customized to match specific experimental scenarios to develop risk-free extreme case lidar datasets. For the sake of this paper, our simulation is modeled to replicate LUMPI dataset (Königsworther Platz intersection, Hannover, Germany).

\noindent \textbf{Details on Sensor Simulation:} Since sensor simulation can highly impact the quality of data, instructions are provided here to achieve best quality. First, the target sensors are identified. This is rather easy, as sensors are usually predetermined for any deployment. The sensors' specifications are noted and converted to simulation scale. For example, simulations in CARLA have 100 units to 1m ratio, that means 100 units of length in CARLA equates to 1 meter in real world. These specification are used to create virtual sensors. The sensor scarification for our employed benchmark dataset, LUMPI, are presented in Table \ref{tab:virtual_sensor_specifications}. Second, the sensors are placed in the HiFi-DT in the exact same pose as of their real counterparts. The pose includes center and height properties of the sensor. For our study, we referred to a single real lidar frame from LUMPI dataset and estimated the positions of sensors using visual queues present in point clouds (e.g., concentric point circles formed by rotating laser reflect from the ground). This practice is feasible as labeling is not required and just a single point cloud is enough to get a good enough approximation of location of sensors. Third, the traffic is generated and the virtual sensors are simulated to get labeled lidar data. For our experimentation, we collected 10K frames using sensor simulation in LUMPI's digital-twin using CARLA simulator.

\begin{table}[ht]
\centering
\caption{Virtual Sensor Specifications}
\label{tab:virtual_sensor_specifications}
\begin{tabular}{|l|c|c|c|c|}
\hline
\multirow{2}{*}{\textbf{Parameter}} & \begin{tabular}[c]{@{}c@{}}\textbf{Hesai}\\ \textbf{Pandar 64}\end{tabular} & \begin{tabular}[c]{@{}c@{}}\textbf{Velodyne}\\ \textbf{HDL-64E}\end{tabular} & \begin{tabular}[c]{@{}c@{}}\textbf{Hesai}\\ \textbf{PandarQT}\end{tabular} & \begin{tabular}[c]{@{}c@{}c@{}}\textbf{Velodyne}\\ \textbf{2 $\times$} \\ \textbf{VLP-16}\end{tabular} \\
\hline
Channels & 64 & 64 & 64 & 16 \\
\hline
Upper FOV ($^\circ$) & 15 & 1.9 & 52.1 & 15 \\
\hline
Lower FOV ($^\circ$) & -25 & -24.6 & -52.1 & -15 \\
\hline
Points per Sec & 2.60M & 4.97M & 0.87M & 0.68M \\
\hline
Max Range (m) & 200 & 120 & 20 & 100 \\
\hline
\end{tabular}
\end{table}

\section{Experiments and Results}

\vspace{1mm}
\noindent \textbf{Training Deep Object Detector:} To test the applicability of the HiFi DT-based simulation, the generated dataset is tested for the 3D object detection task. For this purpose, an off-the-shelf deep learning-based 3D object detector, SEED \cite{liu2024seed}, is employed. Using the HiFi DT-based CARLA simulation, a total of 10,000 labeled frames are generated, compared to benchmark dataset (\textit{Measurement4} subset of the LUMPI dataset \cite{busch2022lumpi}) that contains 8,120 samples. The synthetic dataset was deliberately made slightly larger to highlight that, with adequate scale, high-quality synthetic data can match or even surpass real data in enabling effective Sim2Real learning. The datasets are partitioned into training and testing sets with an 80/20 split, and the model is trained for 12 epochs, using OpenPCDet \cite{openpcdet2020} framework. The hyperparameter configurations are detailed in the Table \ref{tab:hyperparameters}.

\begin{table}[ht]
\centering
\caption{Hyperparameters used in Training of the Deep Object Detector}
\begin{tabular}{|l|l|}
\hline
\textbf{Component} & \textbf{Setting} \\
\hline
Model Size & SEED \cite{liu2024seed} (Large) \\
\hline
Dataset & Training: \texttt{UT-LUMPI} \cite{ucf_ut_lumpi} (synthetic) \\
& Testing: \textit{Measurement4}, LUMPI \cite{busch2022lumpi} (real) \\
\hline
Object Category & Car \\
\hline
Point Cloud Range (m) & $x_{min} = -40.0, x_{max} = 40.0$ \\
(Region of Interest) & $y_{min} = -40.0, y_{max} = 40.0$ \\
& $z_{min} = -8.0, z_{max} = 2.0$ \\
\hline
3D Backbone & VoxelNeXt \cite{chen2023voxelnext} \\
& (5-Stage, 256 Out Channels) \\
\hline
Optimization Algorithm & AdamW with OneCycle LR \\
\hline
Initial Learning Rate & 0.002 \\
\hline
Number of Epochs & 12 \\
\hline
Evaluation Metric & KITTI (AP @ 0.5) \\
\hline
\end{tabular}
\label{tab:hyperparameters}
\end{table}

\vspace{1mm}
\noindent \textbf{Evaluation of Deep Object Detector:} To enable a fair comparison, another SEED model is trained from scratch under identical conditions, utilizing the same hyperparameters and dataset configurations, but on real LUMPI training set instead of synthetic data. This setup ensures that any performance differences can be attributed to the nature of the training data rather than discrepancies in the experimental configuration.

\begin{figure*}[ht]
\centerline{\includegraphics[width=0.98\linewidth]{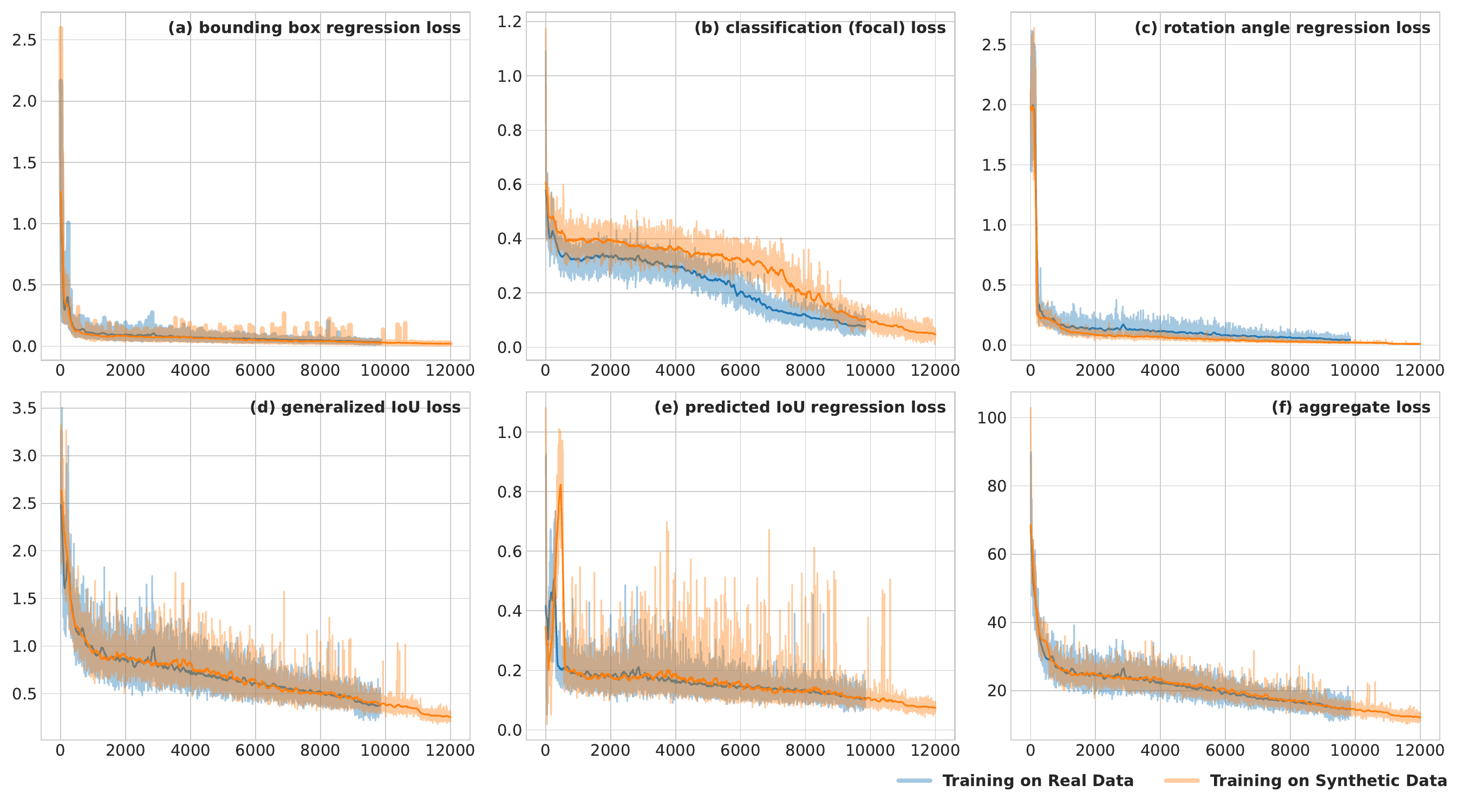}}
\caption{Training loss comparison between real (LUMPI \cite{busch2022lumpi}) and synthetic (UT-LUMPI \cite{ucf_ut_lumpi}) datasets. The plots illustrate the behavior of major loss components of SEED \cite{liu2024seed} during training on the real (blue) dataset and its high-fidelity synthetic counterpart (orange). The loss trends from synthetic data closely follow those from real data, indicating the fidelity of the synthetic dataset. Notably, the synthetic dataset includes slightly more training samples, leading to better convergence behavior that in turn results in superior performance of the model trained on synthetic data.
\label{fig:train_loss}}
\end{figure*}

It is demonstrated in Fig.~\ref{fig:train_loss} that during training loss trends match for both the synthetic and real datasets, suggesting that both datasets share similar underlying distributions relevant to the 3D object detection task. We observe consistent convergence behavior across key loss components. These include, (a) bounding box regression loss that measures $L_1$ difference in box positions and scales, (b) classification loss accounting cross-entropy for object categories, (c)  rotation angle regression loss that evaluates radial distance, (d) generalized IOU (GIOU) loss that penalize box overlap and alignment, and (e) predicted IOU regression loss that is used for confidence score rectification. An (f) aggregate loss trend is also presented to summarize the holistic optimization behavior capturing the combined influence of all individual loss components. A trend line, using rolling mean over 100 steps window around center, is plotted to smooth out fluctuations, enabling clearer comparison between the synthetic and real data training dynamics. The alignment indicates that the synthetic data generated via our HiFi DT pipeline captures task-relevant structural and semantic features comparable to those present in real-world data. Since training loss reflects how well the model is able to extract meaningful patterns from the input data to minimize prediction error, similar loss trajectories imply that the model perceives both domains as statistically and functionally alike. Therefore, we interpret the similarity in loss trends as strong empirical evidence of the representational fidelity and distributional closeness of the synthetic dataset to its real counterpart.

\begin{figure}[h]
\centerline{\includegraphics[width=0.98\linewidth]{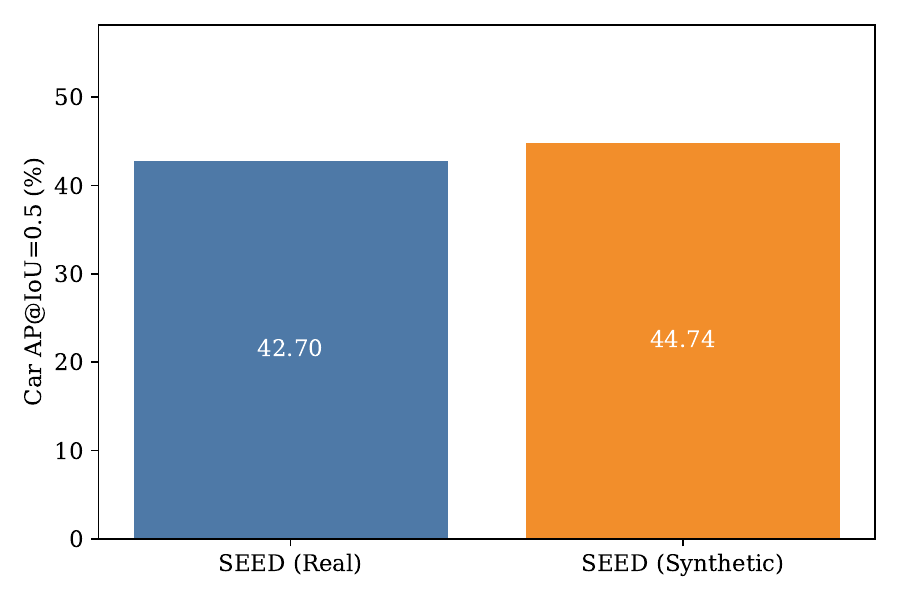}}
\caption{Comparison of detection accuracy of SEED \cite{liu2024seed} models trained on real and synthetic datasets, tested on real dataset. The model trained on synthetic dataset outperforms the identical model trained on real dataset, demonstrating a higher quality of labeling in synthetic dataset.
\label{fig:model_compare}}
\end{figure}

Following the analysis of training loss behavior, we evaluate the detection performance of both models on the real (LUMPI) dataset’s test split. As shown in Figure \ref{fig:model_compare}, the SEED model trained on synthetic data achieves a higher car detection AP@IoU=0.5 (44.74\%) compared to the model trained on real data (42.70\%). This slight performance gain can be attributed to the marginally more training samples available in the synthetic dataset, likely enabling better generalization. Moreover, the superior training loss convergence observed for the synthetic data is reflected in this downstream evaluation, further validating the utility of high-fidelity digital twin-based simulation for real-world 3D object detection.

\vspace{1mm}
\noindent \textbf{Raw-Data and Latent-Feature Analysis:} To empirically estimate the divergence $\Delta(P_{\text{sim}}, P_{\text{real}})$ defined in Section~\ref{subsec:formulation}, we compute multiple geometric and latent-feature alignment metrics including Chamfer Distance (CD), Maximum Mean Discrepancy (MMD), Earth Mover's Distance (EMD), and Fréchet Distance (FD).


Firstly, the statistical similarity is focused. For that, number of points (point density), number of objects (scene complexity), and box sizes (object shapes distribution) are computed for all frames from both the synthetic and the benchmark datasets. Then frame-level means are computed to demonstrate the statistical alignment of the synthetic and real data. These statistics are presented in Fig.~\ref{fig:ut_lumpi_v_lumpi}. As shown in the first panel, both datasets exhibit comparable LiDAR point density, indicating similar spatial coverage. In terms of scene complexity, the average number of 3D bounding boxes per frame is slightly higher in LUMPI (49.3) compared to UT-LUMPI (44.2), but the distributions remain within overlapping variance ranges. The third panel shows the average object volume, with real-world boxes having a mean volume of 12.2 m \textsuperscript{3} versus 10.1 m\textsuperscript{3} in the synthetic set. On right side, the radar plot provides a normalized overview of all metrics, revealing strong alignment between the two domains across point density, object count, and spatial structure.

Secondly, to establish raw structural similarity, Chamfer Distance (CD) is calculated. CD is a commonly used metric that quantifies similarity between two sets of points. It is computed by simply summing the distances between each point in one set and its nearest neighbor in the other set.

Thirdly, for a more in-depth analysis of structural distributions, the latent features of both the synthetic and real datasets are analyzed. Specifically, raw point cloud data is passed through SEED’s backbone network, that extracts high-dimensional latent features rich in contextual information. These features are analyzed through 2D plots, leveraging t-SNE- and UMAP-based dimensionality reduction techniques. Moreover, Maximum Mean Discrepancy (MMD), Earth Mover's Distance (EMD), and Fréchet Distance (FD) are also computed to quantify the alignment between real and synthetic data.

\begin{figure*}[t]
\centering
\begin{minipage}{\linewidth}
    \centering
    \includegraphics[width=0.98\linewidth]{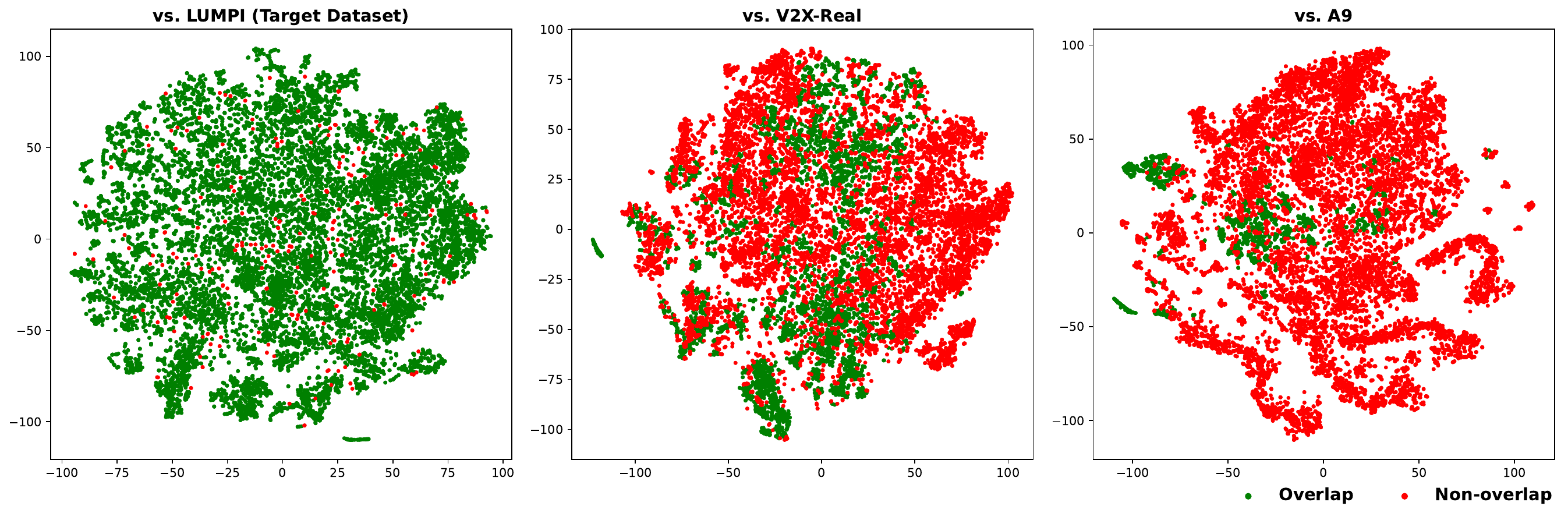}
    \caption{t-SNE visualization of SEED backbone features comparing the synthetic high-fidelity digital-twin-based dataset (UT-LUMPI) to target (LUMPI) and two other real-world datasets. Each sub-plot shows the 2D t-SNE projection of features extracted from the SEED backbone for UT-LUMPI against a real dataset: LUMPI (left), V2X-Real (center), and A9 (right). Green points represent overlapping (similar) regions, while red points indicate non-overlapping feature regions. The high degree of overlap with the LUMPI target dataset suggests strong distributional alignment emerging due to digital-twin modeling approach, while significantly lower overlap with V2X-Real and A9 indicates domain mismatch.}
    \label{fig:tnse_all_overlap}
\end{minipage}

\vspace{1em} 

\begin{minipage}{\linewidth}
    \centering
    \includegraphics[width=0.98\linewidth]{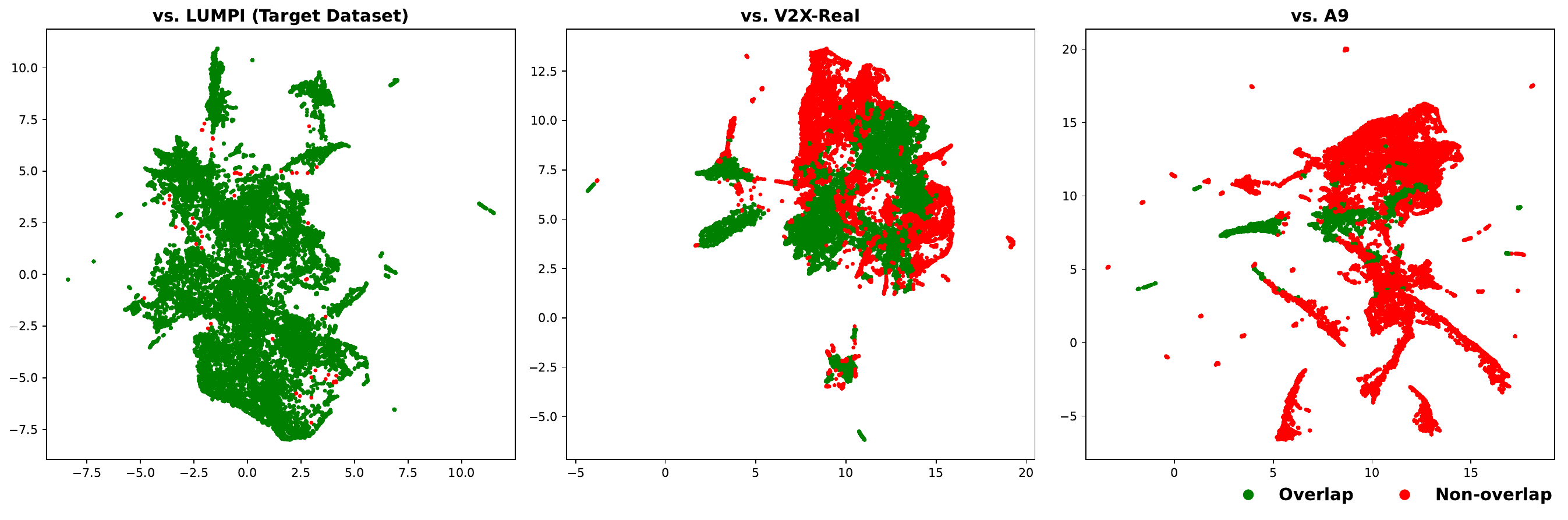}
    \caption{UMAP-based comparison of feature distributions between the HiFi-DT-based dataset and three real-world datasets: LUMPI (target location for HiFi-DT modeling), V2X-Real, and A9. Similar to Fig.~\ref{fig:tnse_all_overlap}, the concentrated alignment with LUMPI underscores the representational fidelity of the synthetic data, while reduced overlap with V2X-Real and A9 illustrates the domain divergence commonly encountered across unrelated datasets.}
    \label{fig:umap_all_overlap}
\end{minipage}
\end{figure*}

t-SNE (t-distributed Stochastic Neighbor Embedding) and UMAP (Uniform Manifold Approximation and Projection) are employed in this study as nonlinear dimensionality reduction techniques to visualize and analyze the high-dimensional latent feature space extracted from the point cloud data. t-SNE focuses on preserving local neighborhood structures, making it particularly effective at capturing fine-grained relationships between data points, whereas UMAP emphasizes both local and global structure preservation, providing a balance between detailed clustering and broader data distribution patterns. By projecting the high-dimensional features into two-dimensional space, these methods facilitate qualitative assessment of the structural and contextual alignment between the synthetic and real datasets. In Fig.~\ref{fig:tnse_all_overlap}, t-SNE projections of the SEED backbone features for the HiFi-DT-based synthetic dataset (UT-LUMPI) against real datasets are presented. These dataset include LUMPI (target), V2X-Real, and A9. The HiFi-DT based synthetic dataset features exhibit strong overlap to target location data in the embedding space, with the majority of synthetic and real points occupying similar regions. Such high overlap presents the evidence of the high distributional similarity achieved by the HiFi-DT approach, enabling the model trained on synthetic data to generalize well to real data. In contrast, the limited overlap with V2X-Real and A9 highlights the domain shift that typically exists across different datasets, even when captured under seemingly similar roadside conditions. While there are domain adaptation techniques that attempt to mitigate such mismatches, the HiFi-DT method offers a straightforward and scalable way to synthesize data that is naturally aligned with the target deployment environment.



\noindent Additionally, Uniform Manifold Approximation and Projection for Dimension Reduction (UMAP) comparison of SEED backbone is presented in Fig.~\ref{fig:umap_all_overlap}. UMAP is better in preserving both local and global structures. Similar to Fig.~\ref{fig:tnse_all_overlap}, latent features of HiFi-DT-based data exhibit dense and cohesive overlap with the LUMPI dataset, again reflecting strong distributional alignment achieved by our approach. In contrast, feature overlap with V2X-Real and A9 remains sparse, revealing significant domain discrepancies that challenge transfer learning across datasets.

\begin{figure*}[ht]
\centerline{\includegraphics[width=0.98\linewidth]{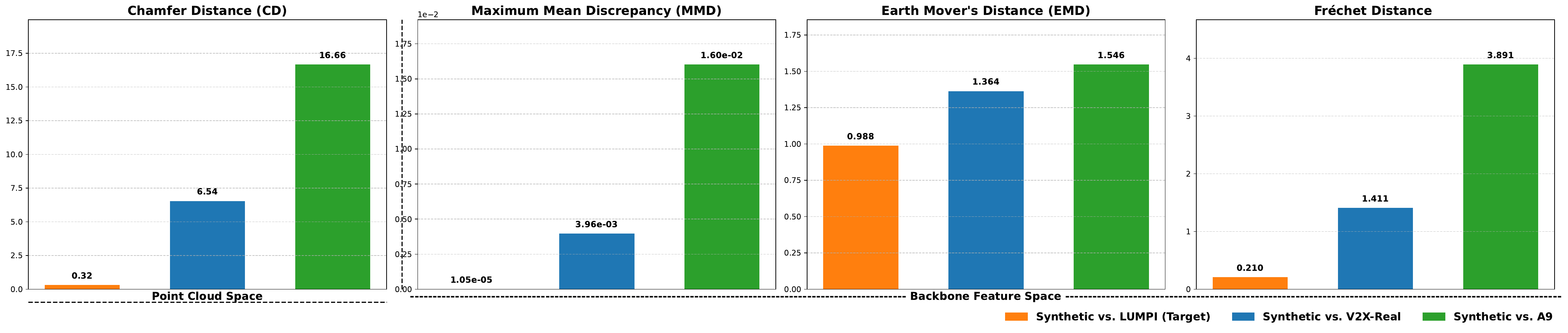}}
\caption{Quantitative comparison of distributional similarity between HiFi-DT-based and real data features using geometric and latent-feature metrics. The figure reports four metrics—Chamfer Distance (CD), Maximum Mean Discrepancy (MMD), Earth Mover’s Distance (EMD), and Fréchet Distance—across two spaces: point cloud geometry and SEED backbone latent-feature embeddings. Comparisons are made between synthetic data and three real data features: LUMPI (target), V2X-Real, and A9. Across all metrics and both spaces, UT-LUMPI shows the closest alignment with its target, LUMPI, validating its suitability for domain-specific training.
\label{fig:cd_emd_mmd_fd}}
\end{figure*}

For the quantitative analysis of distributional similarity of HiFI-DT-based data, we compute a suite of standard domain alignment metrics across both raw point cloud and deep feature spaces. First, Chamfer Distance (CD) is reported to evaluate the geometric similarity between point clouds in the original spatial domain. Unlike the latent feature space metrics, CD directly measures the average nearest-neighbor distance between two point sets, offering first quantitative measure on how closely the point clouds align. Maximum Mean Discrepancy (MMD) and Earth Mover's Distance (EMD) are utilized to assess the deep latent feature space. MMD is a kernel-based statistical test that measures the distance between the means of two distributions in a reproducing kernel Hilbert space, offering a rigorous indication of how closely the two datasets align at a distributional level. EMD, also known as the Wasserstein distance, quantifies the minimal cost required to transform one distribution into another, capturing both local and global differences between data distributions. Additionally, Fréchet Distance (FD) compares the mean and covariance of the two distributions in the feature space for evaluating generative quality in embedding domains. MMD is sensitive to differences in higher-order statistics, EMD captures the geometric structure of the distributions, and FD summarizes both first- and second-order distributional characteristics—thus enabling a comprehensive evaluation of the synthetic data's fidelity to the real data. The CD, MMD, EMD, and FD scores are presented in Fig.~\ref{fig:cd_emd_mmd_fd}.

HiFi-DT-based data consistently exhibits the lowest distance to its target dataset (LUMPI) across all four measures—Chamfer Distance (0.32), MMD (1.05e-5), EMD (0.988), and Fréchet Distance (0.210). In contrast, significantly higher distances are observed when comparing the synthetic data to unrelated datasets such as V2X-Real and A9. These results confirm that the HiFi digital twin methodology produces data that is not only visually aligned but also statistically close in both geometry and learned representation spaces. 

\section{Conclusion}
This paper presented a systematic methodology for bridging the Sim2Real learning gap in lidar-based perception tasks through the development and deployment of high-fidelity digital twins (HiFi DTs). By integrating realistic static scene geometry, lane-level road topology, dynamic traffic distributions, and precise sensor modeling, the proposed approach enables the generation of synthetic lidar datasets that are structurally and statistically aligned with real-world data. Experimental evaluations on a standard 3D object detection task demonstrated that models trained solely on HiFi-DT-based synthetic data not only achieve competitive performance but also surpass the accuracy of models trained on real data by 4.8\%.

Furthermore, detailed statistical and structural analyses at both raw-input and latent-feature levels, utilizing metrics such as Chamfer Distance, Maximum Mean Discrepancy (MMD), Earth Mover’s Distance (EMD), t-SNE, and UMAP, substantiated the distributional alignment of synthetic and real data. These findings highlight the efficacy of HiFi DTs in mitigating the domain shift that traditionally hampers the transferability of simulation-trained models to real-world environments.

Unlike traditional domain adaptation techniques that require post-hoc corrections or fine-tuning, the proposed HiFi DT framework inherently generates in-domain data, obviating the need for complex adaptation pipelines. This enables rapid, scalable, and cost-effective development of lidar datasets tailored to specific locations and scenarios, including rare and extreme cases that are otherwise difficult to capture.

This work underscored the pivotal role of high-fidelity digital twins in advancing simulation-based training paradigms for intelligent transportation systems and offers a practical pathway for generating high-quality, task-relevant synthetic data. Future work will extend this framework to support a broader range of perception tasks, incorporate larger DTs and more sensor models for autonomous driving applications, and explore fully automated pipelines for digital twin construction from publicly available geospatial data.

\printbibliography

\newpage

\begin{IEEEbiography}[{\includegraphics[width=1in,height=1.25in,clip,keepaspectratio]{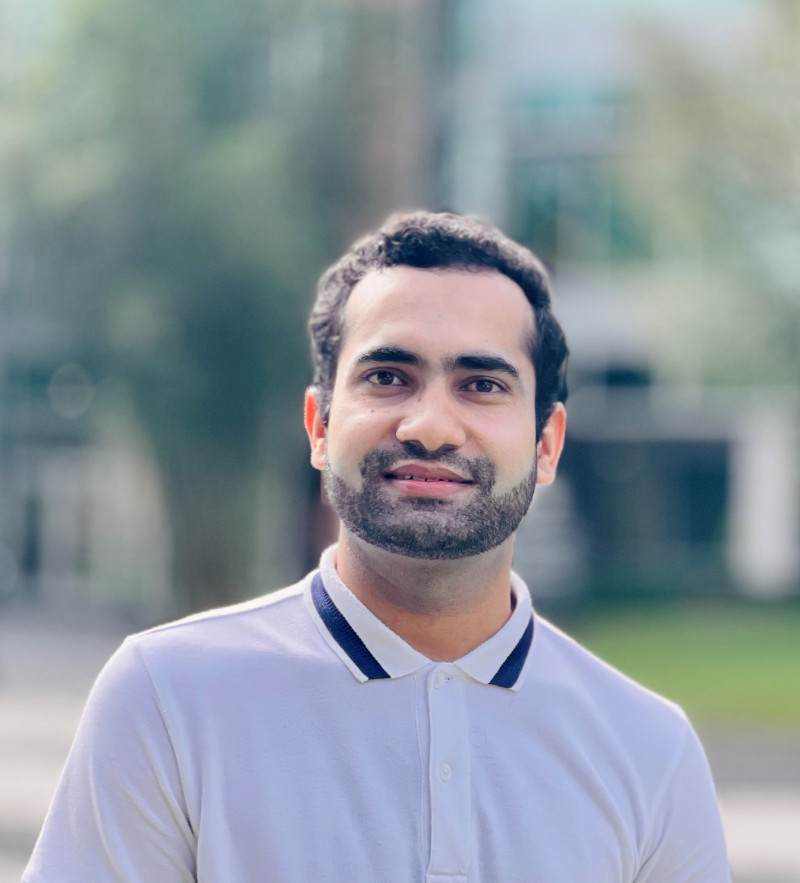}}]{Muhammad Shahbaz } is passionate about interdisciplinary research across Advanced Computer Vision and AI and their applications in the field of intelligent transportation systems and intelligent robotics. He received the B.S. in computer science degree from Pir Mehr Ali Shah Arid Agriculture University Rawalpindi, and M.S. degree in computer science from the Pakistan Institute of Engineering and Applied Sciences, Islamabad, Pakistan. Since 2021, he is pursuing Ph.D. in Civil Engineering at University of Central Florida, USA. His primary focus during Ph.D. spanned efficient and effective 3D perceptions systems for Intelligent Transportation Systems using high-fidelity simulation and multi-modal sensor fusion.
\end{IEEEbiography}

\begin{IEEEbiography}[{\includegraphics[width=1in,height=1.25in,clip,keepaspectratio]{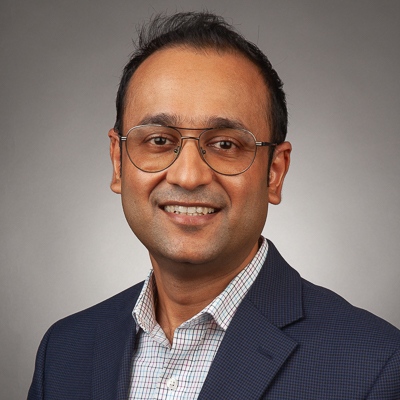}}]
{SHAURYA AGARWAL } 
 (Senior Member, IEEE) is currently an Associate Professor in the Civil, Environmental, and Construction Engineering Department at the University of Central Florida. He is the founding director of the Urban Intelligence and Smart City (URBANITY) Lab, Director of the Future City Initiative at UCF, and serves as the coordinator for Smart Cities Masters program at UCF. He was previously (2016-18) an Assistant Professor in the Electrical and Computer Engineering Department at California State University, Los Angeles. He completed his post-doctoral research at New York University (2016) and his Ph.D. in Electrical Engineering from the University of Nevada, Las Vegas (2015). His B.Tech. degree is in ECE from the Indian Institute of Technology (IIT), Guwahati. His research focuses on interdisciplinary areas of cyber-physical systems, smart and connected transportation, and connected and autonomous vehicles. Passionate about cross-disciplinary research, he integrates control theory, information science, data-driven techniques, and mathematical modeling in his work. As of May 2025, he has published a book, over 37 peer-reviewed journal publications, and multiple conference papers. His work has been funded by several private and government agencies. He is a \textit{senior member} of IEEE and serves as an \textit{Associate Editor} of IEEE Transactions on Intelligent Transportation Systems.
\end{IEEEbiography}

\vfill

\end{document}